\documentclass[english,12pt,a4paper, bibtotoc]{article}

\usepackage[pdftex,
    bookmarks,
    bookmarksopen=true,
    bookmarksnumbered=true,
    pdfauthor={D.~P.~Mohr, P.~Huber, M.~Schwabe, C.~A.~Knapek},
    pdftitle={3D Particle Positions from Computer Stereo Vision in PK-4},
    pdfkeywords={},
    colorlinks, linkcolor=blue, urlcolor=blue
]{hyperref}

\usepackage{graphicx}
\usepackage{color}

\usepackage{babel}
\usepackage{csquotes}
\MakeOuterQuote{"}

\usepackage{multirow}

\usepackage{amssymb}
\usepackage{amsmath}
\usepackage{natbib}
\usepackage{subcaption}
\usepackage[all]{xy}
\newcommand{\R}{\ensuremath{\mathbb{R}}}
\newcommand{\norm}[1]{\left|\left|#1\right|\right|}

\newcommand{\figref}[1]{\autoref{#1}}
\newcommand{\tabref}[1]{\autoref{#1}}
\newcommand{\secref}[1]{\autoref{#1}}

\title{3D Particle Positions from Computer Stereo Vision in PK-4}
\author{\mbox{Daniel P. Mohr}, \mbox{Peter Huber}, \mbox{Mierk Schwabe}, \mbox{Christina A. Knapek}}
\date{\today}
\begin{document}
\maketitle

\tableofcontents

\section{Introduction} %%%%%%%%%%%%%%%%%%%%%%%

\begin{figure}\center
  \setlength{\unitlength}{1mm}
  \begin{picture}(110,80)(0,0)
    \put(0,0){\includegraphics[height=80mm,trim=27mm 12mm 27mm 12mm,clip]{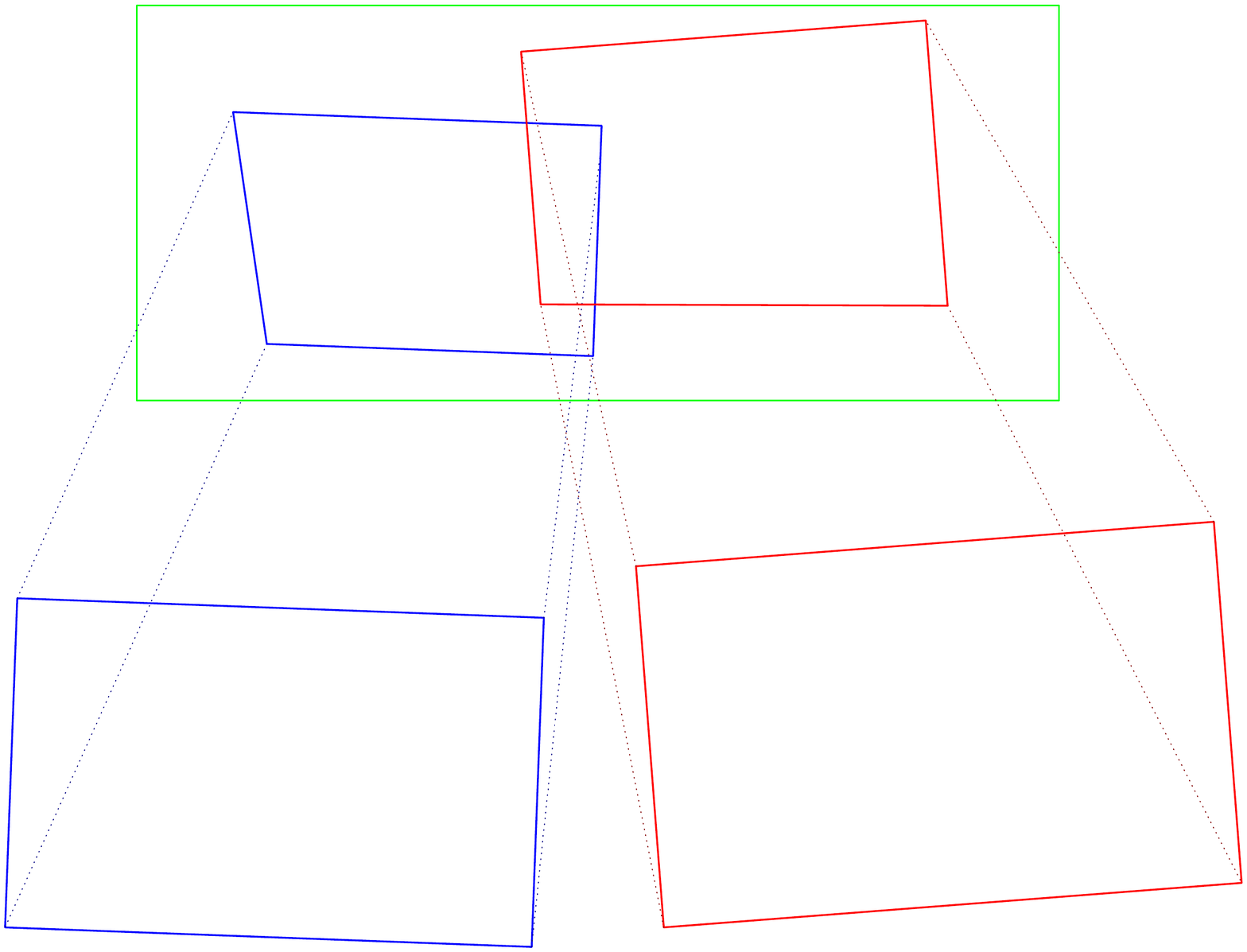}}
    \put(3,3){\color{blue}image plane cam$1$}
    \put(58,6){\color{red}image plane cam$2$}
    \put(12,77){\color{green}laser sheet/plane}
    \put(22,66){\color{blue}FoV cam$1$}
    \put(55,72){\color{red}FoV cam$2$}
  \end{picture}
  \caption{General camera setup of the PK-4 experimental apparatus. This is only a sketch to show the general concept -- e.\ g. scale, angles does not reflect the real situation.}\label{fig:setup}
\end{figure}

Complex plasmas consist of microparticles embedded in a low-temperature plasma containing ions, electrons and neutral particles. The microparticles form a dynamical system that can be used to study a multitude of effects on the level of the constituent particles \citep{Morfill2009}. The microparticles are usually illuminated with a sheet of laser light, and the scattered light can be observed with digital cameras --- see \figref{fig:setup}. \citet{mohr:2019} describe the feature (particle) detection in a single image/frame.
Some complex plasma microgravity research facilities \citep{Pustylnik:2016} use two cameras with an overlapping field of view. \citet{Zaehringer:2018} describes this for a parabolic flight setup of Ekoplasma. Other facilities use even more cameras \citep{Himpel2011}. 

An overlapping field of view can be used to combine the resulting images into one and trace the particles in the larger field of view. In previous work this was discussed for the images recorded by the PK-4 Laboratory on board the International Space Station \citep{Schwabe:2019}. In that work the width of the laser sheet was, however, not taken into account. In this paper, we will discuss how to improve the transformation of the features into a joint coordinate system, and possibly extract information on the 3D position of particles in the overlap region.

\section{Modeling} %%%%%%%%%%%%%%%%%%%%%%%
In this section the standard model for computer stereo vision \citet{Hartley:2004} is briefly touched in a scope necessary for this document. This model is widely used and is, for example, available as an open source implementation \citet{opencv_library}.

The standard model for computer stereo vision uses the pinhole camera model: We can describe the mapping from world coordinates $x \in \R^3$ to image coordinates $p \in \R^2$ up to a scaling parameter $s$:

\begin{align*}
  s \left(\begin{array}{c}
    p \\
    1 \\
  \end{array}\right) = \underbrace{C_i \left(\begin{array}{c|c}
      R_i & t_i \\
    \end{array}\right)}_{=: P_i} \left(\begin{array}{c}
    x \\
    1 \\
  \end{array}\right)
\end{align*}

with
\begin{align*}
  C_i \in \R^{3 \times 3} & : \mbox{ camera matrix of camera $i$}\\
  R_i \in \R^{3 \times 3} & : \mbox{ rotation matrix of camera $i$}\\
  t_i \in \R^{3} & : \mbox{ translation vector of camera $i$}\\
  P_i \in \R^{3 \times 4} & : \mbox{ projection matrix of camera $i$.}\\
\end{align*}

In addition we use the rational distortion model, which is also used by opencv \citep{opencv_library, Claus:2005}.

\subsection{Possible Accuracy in View Direction} \label{sec:Possible Accuracy in View Direction} %%%%%%%%%%%%%%%%%%%%%%%
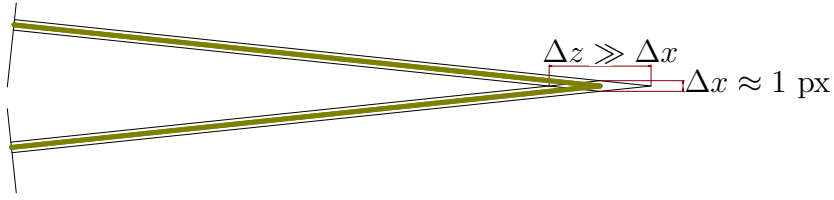
\begin{figure}\center
  \setlength{\unitlength}{1cm}
  \linethickness{0.5mm}
  \begin{picture}(12, 3)
    \put(0,0){
      \includegraphics[width=9.73cm,trim={0 6cm 0 6cm},clip]{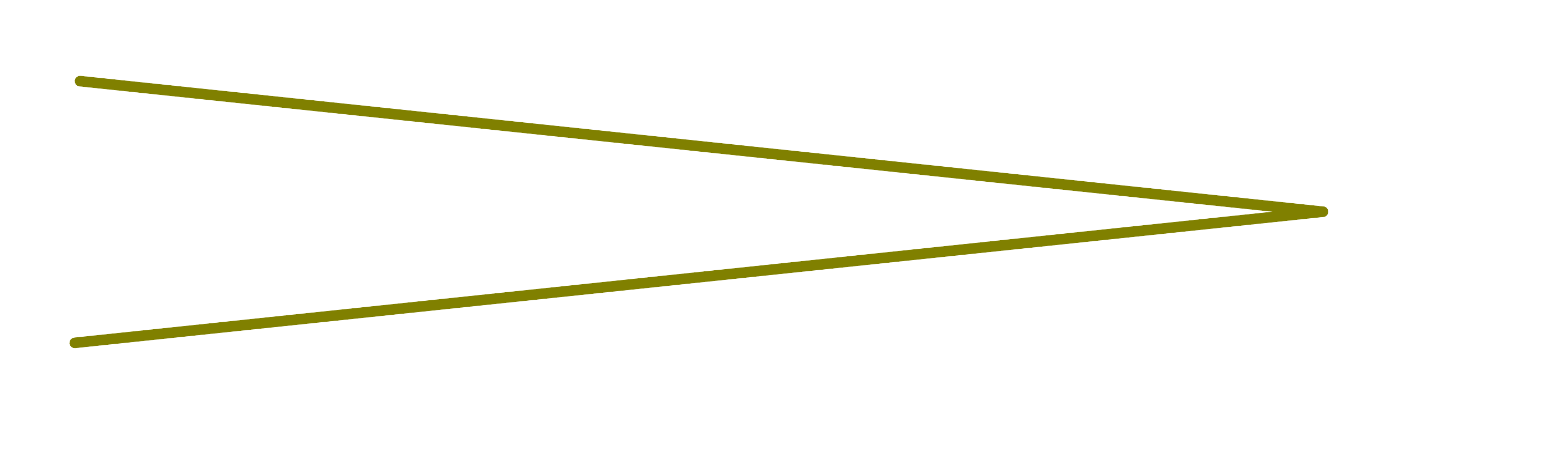}
    }
    \put(0,0){
      \includegraphics[width=9.73cm,trim={0 6cm 0 6cm},clip]{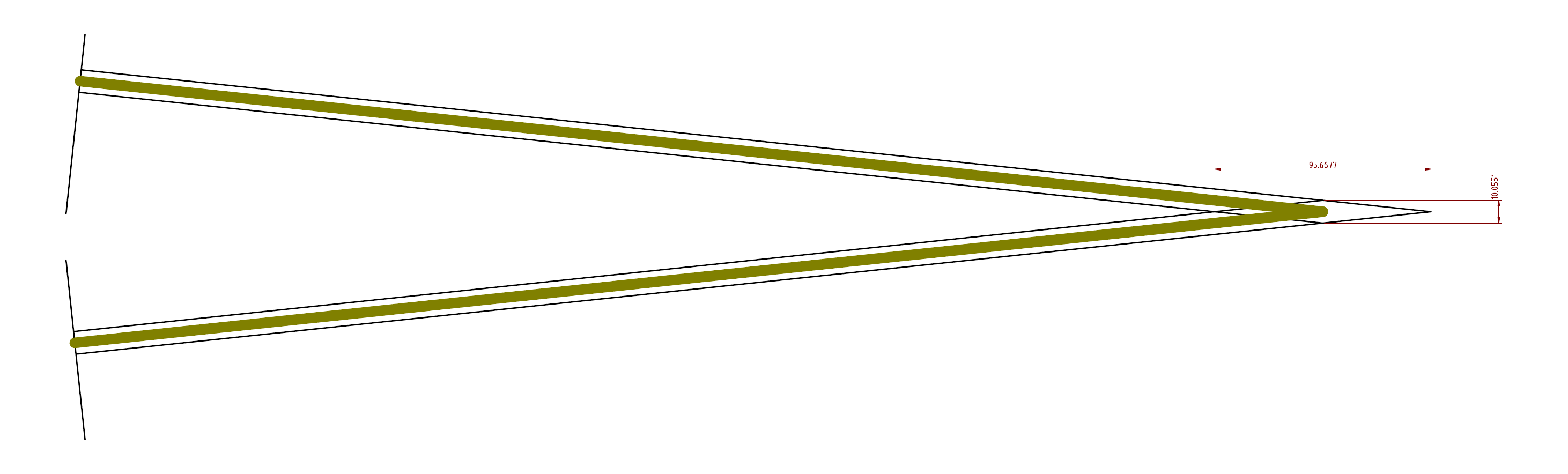}
    }
    \put(8.2, 1.92){\color{white}
      \line(1,0){0.3}
    }
    \put(7.45, 1.95){
      $\Delta z \gg \Delta x$
    }
    \put(9.2, 1.73){\color{white}
      \line(1,0){0.3}
    }
    \put(9.2, 1.77){\color{white}
      \line(1,0){0.3}
    }
    \put(9.2, 1.81){\color{white}
      \line(1,0){0.3}
    }
    \put(9.2, 1.85){\color{white}
      \line(1,0){0.3}
    }
    \put(9.32, 1.55){
      $\Delta x \approx 1 \mbox{ px}$
    }
  \end{picture}
  \caption{View of two cameras with a small angle of $\alpha$.}\label{fig:accuracy}
\end{figure}

If we consider a feature that is represented by a single pixel in each camera --- see \figref{fig:accuracy} --- we see that the intersection is a parallelogram.
The lengths of the diagonals of the parallelogram, $\Delta z$ and $\Delta x$, in relation to the height of the feature, $h = 1 \mbox{ px}$, are:
\[ \sqrt{\left(\frac{h}{\sin{\beta}}\right)^2 + \left(\frac{h}{\sin{\alpha}}\right)^2 \pm 2 \frac{h}{\sin{\beta}} \frac{h}{\sin{\alpha}} \cos{\alpha}}, \]
%import numpy
%alpha = numpy.pi * 6.3 / 180.0 # (Blickwinkel/Öffnungswinkel)
%alpha = numpy.pi * 11 / 180.0 # (Blickwinkel/Öffnungswinkel)
%beta = numpy.pi - alpha
%h_a = 1 # px
%h_b = 1 # px
%b = h_a / numpy.sin(alpha)
%a = h_b / numpy.sin(beta)
%numpy.sqrt(a**2 + b**2 + 2 * a * b * numpy.cos(alpha))

where $\alpha$ is the angle of view and $\beta = \pi - \alpha$.

The PK-4 setup uses the camera Basler Pilot piA1600-35gm \citep{Pustylnik:2016} with a chip of $11.9 \mbox{ mm} \times 8.9 \mbox{ mm}$) and a $50 \mbox{ mm}$ objective lens. The field of view is about $22.4 \mbox{ mm} \times 16.8 \mbox{ mm}$. In the overlap we get:
\begin{itemize}
\item $\alpha = 11^\circ$
\item $\Delta z \approx 10 \mbox{ px} \gg 1 \mbox{ px} \approx \Delta x$
\end{itemize}

\section{Finding Corresponding Features}
Since the images of the two cameras of the PK-4 setup are very similar in the overlapping region, we can identify corresponding features just by looking at the same position. Therefore we need to combine the images from the cameras.

Using a clipping of one image, we can compare it to any possible clipping of the same size of the other image. The similarity measure ''Normalized Cosine Similarity`` of the vectors a $\in \R^n$ and $b \in \R^n$ is given by:

\[ \frac{1}{2} \left(1 + \frac{a \cdot b}{\norm{a}_2 \norm{b}_2}\right) \in [0, 1] \]

\begin{figure}\center
  \begin{subfigure}[b]{0.9\textwidth}
    \includegraphics[width=\textwidth]{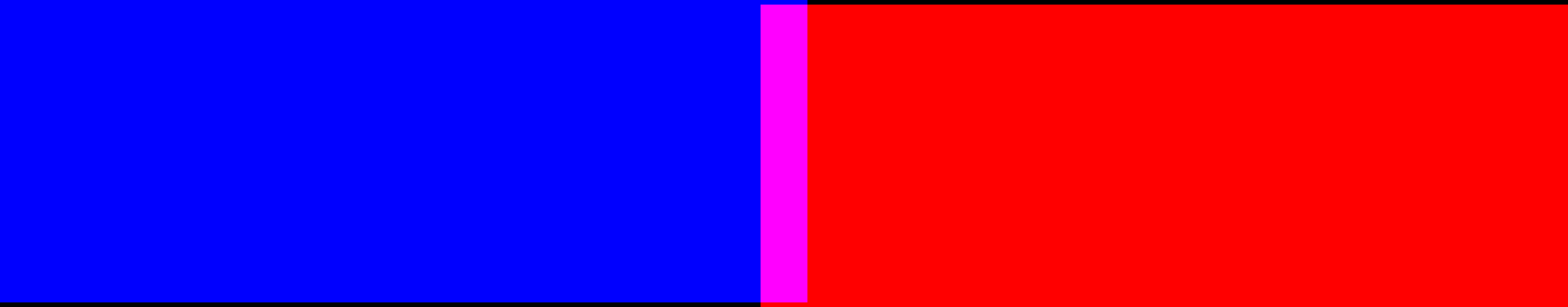}
    \caption{overview of the matching}
  \end{subfigure}\\
  \begin{subfigure}[b]{0.9\textwidth}
    \includegraphics[width=\textwidth]{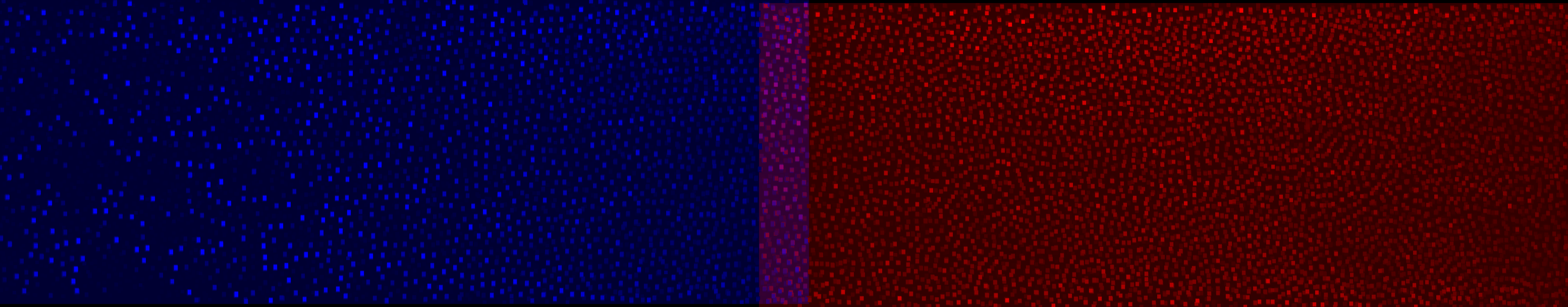}
    \caption{matching}\label{fig:matching}
  \end{subfigure}\\
  \begin{subfigure}[t]{0.3\textwidth}
    \includegraphics[width=\textwidth, trim={1503 508 1503 0}, clip]{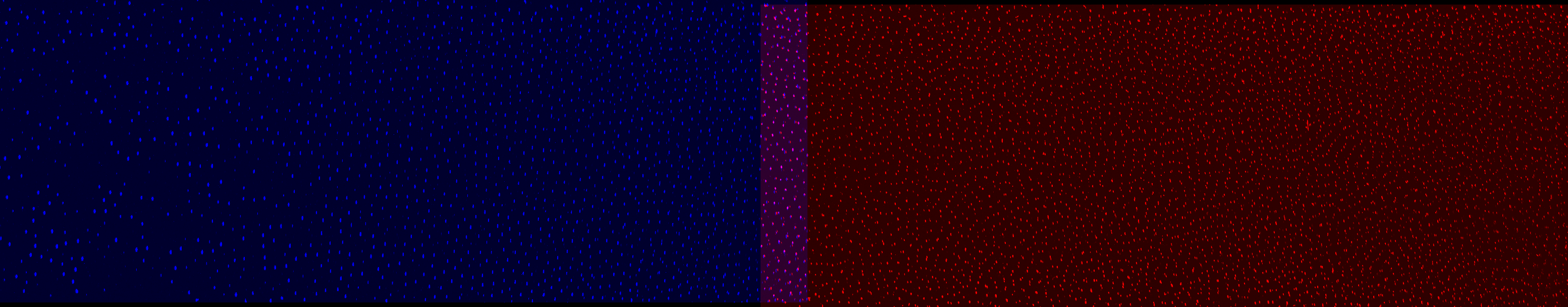}
    \caption{clipping of \figref{fig:matching} of the center middle part}
  \end{subfigure}
  \begin{subfigure}[t]{0.3\textwidth}
    \includegraphics[width=\textwidth, trim={1503 254 1503 254}, clip]{figures/pk4_defocused_laser_combine_simple_brighter.png}
    \caption{clipping of \figref{fig:matching} of the top middle part}
  \end{subfigure}
  \begin{subfigure}[t]{0.3\textwidth}
    \includegraphics[width=\textwidth, trim={1503 0 1503 508}, clip]{figures/pk4_defocused_laser_combine_simple_brighter.png}
    \caption{clipping of \figref{fig:matching} of the bottom middle part}
  \end{subfigure}
  \caption{Two images are combined using the common technique template matching. This means only translations are used to embed the right (red) image in the left (blue) one.}\label{fig:template matching}
\end{figure}
The clipping with the highest similarity allows to combine the image in a simple way as shown in \figref{fig:template matching}. In this combined image \figref{fig:matching} it is easy to identify corresponding features, which then form the basis for the further analysis.

\section{Projection to Laser}
Typically in computer stereo vision the images of cameras are used to reconstruct the real scene. In complex plasmas only a small area of the real world is illuminated by a laser. Taken this fact into account allows to reduce the epipolar lines from infinitely long lines to short line segments.

\subsection{Projection to Laser Plane}
In the intersection of the field of views ($F_{oV}$) of the cameras, the classical approach of a pinhole camera model leads to
\begin{align*}
  s_i \left(\begin{array}{c}
    p_j^{(i)} \\
    1 \\
  \end{array}\right) = P_i \left(\begin{array}{c}
    x_j \\
    1 \\
  \end{array}\right) \mbox{ for all } x_j \in F_{oV} \subset \R^3
\end{align*}

Typically the path of a particle is traced in one image plane. This is only meaningful with the assumption that the laser sheet has no depth.

Here, $F_{oV}$ determines the intersection of the field of views of the cameras.

The $F_{oV}$ has to be in the laser plane:
\begin{align*}
  \forall x_j \in F_{oV}: l^T x_j = 0.
\end{align*}

In \citep{Pustylnik:2016,Zaehringer:2018} the projection matrices $P_1$ and $P_2$ are not known. We have to identify them from the overlapping fields of view.

Let us assume we have corresponding features $p_i^{(j)}$ by identifying the features in the combined image --- \figref{fig:template matching}. Now we can write for all features:

\begin{align*}
  \left(\begin{array}{c}
    p_j^{(1)} \\
    1 \\
    p_j^{(2)} \\
    1 \\
    0 \\
  \end{array}\right) = \left(\begin{array}{c|c}
    \multicolumn{2}{c}{\frac{1}{s_1} C_1 \left(\begin{array}{c|c}
      R_1 & t_1 \\
    \end{array}\right)} \\\hline
    \multicolumn{2}{c}{\frac{1}{s_2} C_2 \left(\begin{array}{c|c}
      R_2 & t_2 \\
    \end{array}\right)} \\\hline
    l^T & 0 \\
  \end{array}\right) \left(\begin{array}{c}
    x_j \\
    1 \\
  \end{array}\right) \in \R^7
\end{align*}

This looks like a linear equation, but in the unknowns $f_x^{(i)}$, $f_y^{(i)}$, $c_x^{(i)}$, $c_y^{(i)}$, $\alpha_i$, $\beta_i$, $\gamma_i$, $t_i$ and $x_j$ this is nonlinear due to the trigonometric functions in the rotation matrices. Furthermore, the distortion model (not formulated here) is not linear.

Without loss of generality we can choose a convenient coordinate system and thus we get for the laser plane:
\begin{align*}
  \forall x_j \in F_{oV}: x_j^T \left(\begin{array}{c} 0 \\ 0 \\ 1 \\ \end{array}\right) = 0.
\end{align*}

If we assume to have $n \geq 5$ corresponding pairs of features, this leads to an overdetermined system of equations, and we have to find the best solution:

\begin{align*}
  \mbox{obj. func.: } & \sum_{j=1}^{n}{\norm{
  \left(\begin{array}{c}
    p_j^{(1)} \\
    1 \\
    p_j^{(2)} \\
    1 \\
  \end{array}\right) - \left(\begin{array}{cccc}
    \multicolumn{4}{c}{\frac{1}{s_1} C_1 \left(\begin{array}{c|c}
      R_1 & t_1 \\
    \end{array}\right)} \\\hline
    \multicolumn{4}{c}{\frac{1}{s_2} C_2 \left(\begin{array}{c|c}
      R_2 & t_2 \\
      \end{array}\right)}\\
  \end{array}\right) \left(\begin{array}{c}
    x_j \\
    1 \\
  \end{array}\right)
    }^2} \rightarrow \min_{s_i, C_i, R_i, t_i, x_j}\\
  \mbox{ s.t. } & x_j^T \left(\begin{array}{c} 0 \\ 0 \\ 1 \\ \end{array}\right) = 0
\end{align*}
\begin{figure}\center
  \begin{subfigure}[t]{0.7\textwidth}
    \includegraphics[width=\textwidth, trim={5cm 3cm 5cm 4cm}, clip]{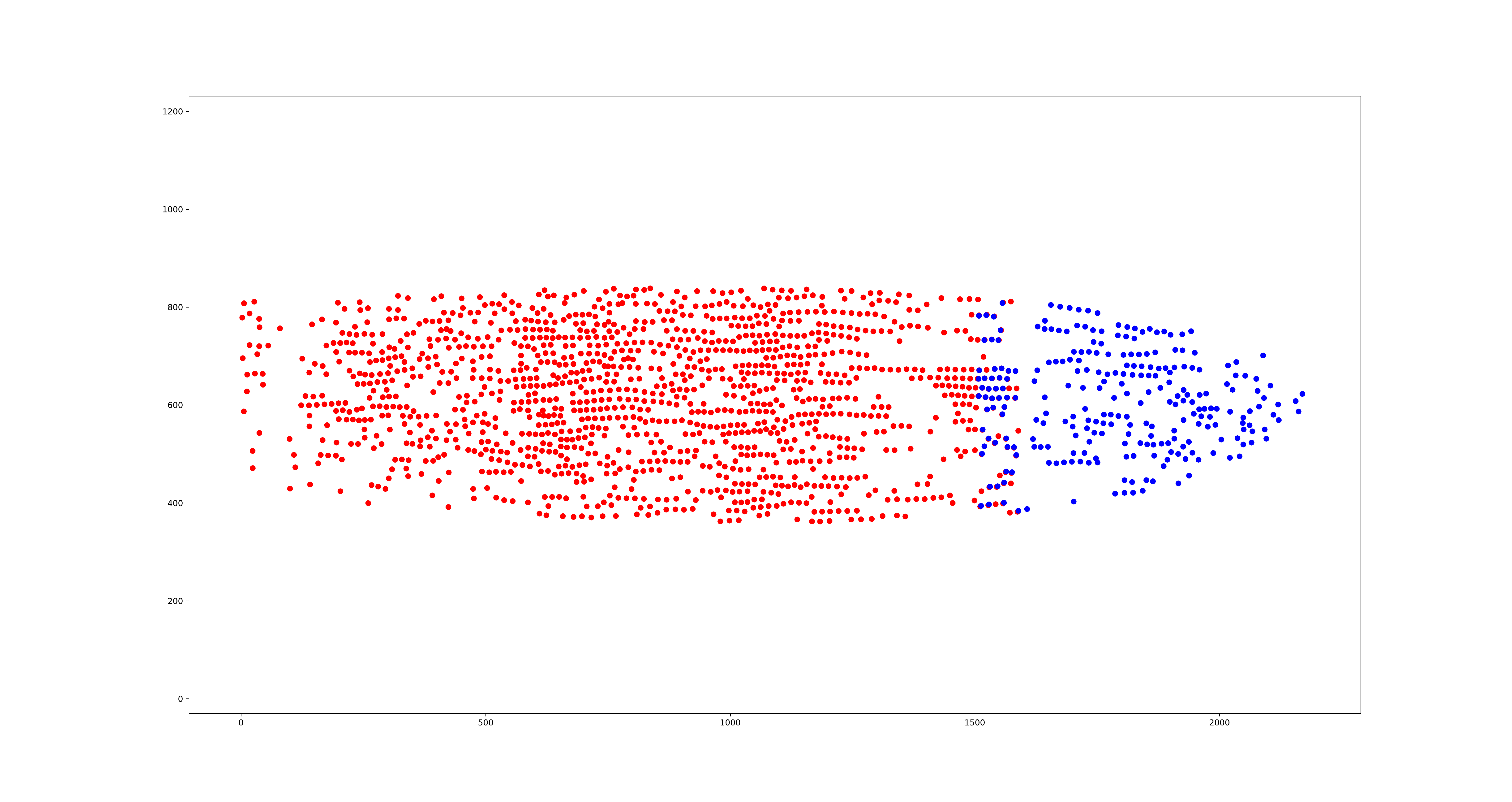} % 650 mm x 345 mm
    \caption{simple combined features}\label{fig:focused laser simple combined}
  \end{subfigure}
  \begin{subfigure}[t]{0.25\textwidth}
    \includegraphics[width=\textwidth, trim={40.9cm 18cm 21.1cm 12.5cm}, clip]{figures/combine_images_simple_example_2.pdf}
    \caption{clipping of \figref{fig:focused laser simple combined} of the top overlapping region}
  \end{subfigure}\\
  \begin{subfigure}[t]{0.7\textwidth}
    \includegraphics[width=\textwidth, trim={5cm 3cm 5cm 4cm}, clip]{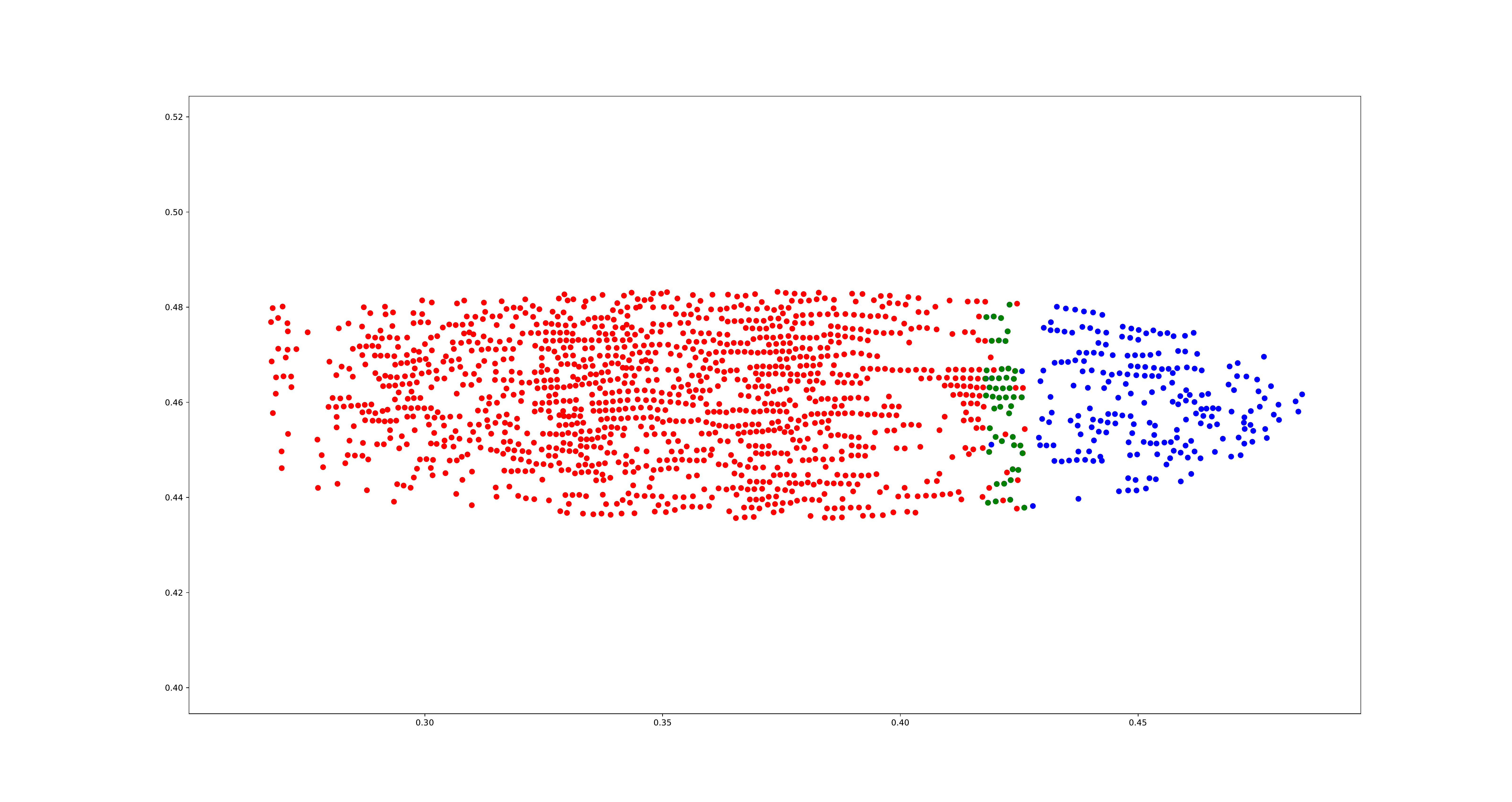} % 650 mm x 345 mm
    \caption{combined features}\label{fig:focused laser combined}
  \end{subfigure}
  \begin{subfigure}[t]{0.25\textwidth}
    \includegraphics[width=\textwidth, trim={41.25cm 18cm 20.75cm 12.5cm}, clip]{figures/combine_images_example_2.pdf}
    \caption{clipping of \figref{fig:focused laser combined} of the top overlapping region}
  \end{subfigure}
  \caption{Projection to laser plane: Example with focused laser}\label{fig:projection laser plane focused laser}
\end{figure}
\begin{figure}\center
  \begin{subfigure}[t]{0.7\textwidth}
    \includegraphics[width=\textwidth, trim={5cm 3cm 5cm 4cm}, clip]{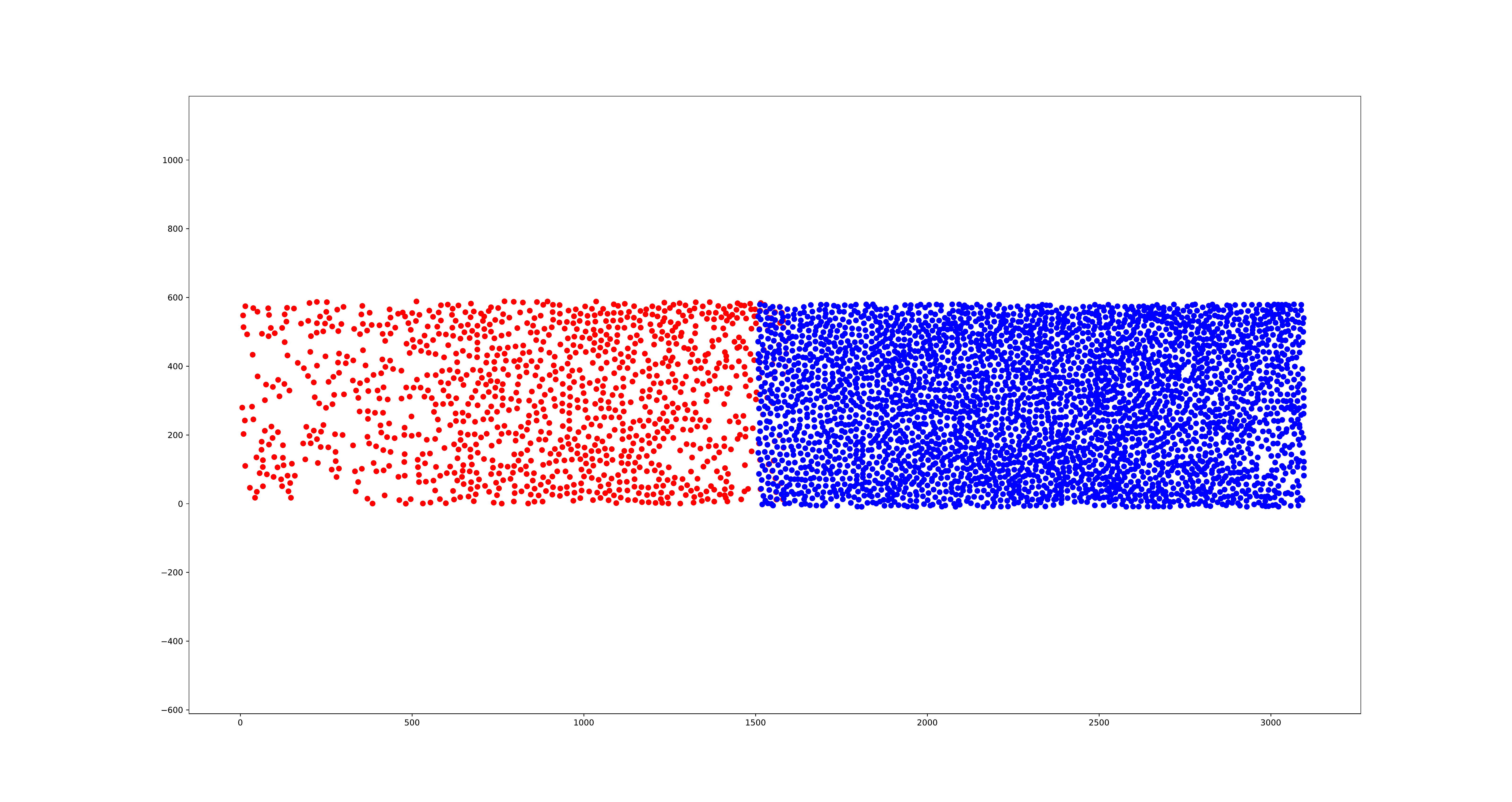} % 650 mm x 345 mm
    \caption{simple combined features}\label{fig:defocused laser simple combined}
  \end{subfigure}
  \begin{subfigure}[t]{0.29\textwidth}
    \includegraphics[width=\textwidth, trim={2.5cm 2.5cm 2cm 2.5cm}, clip]{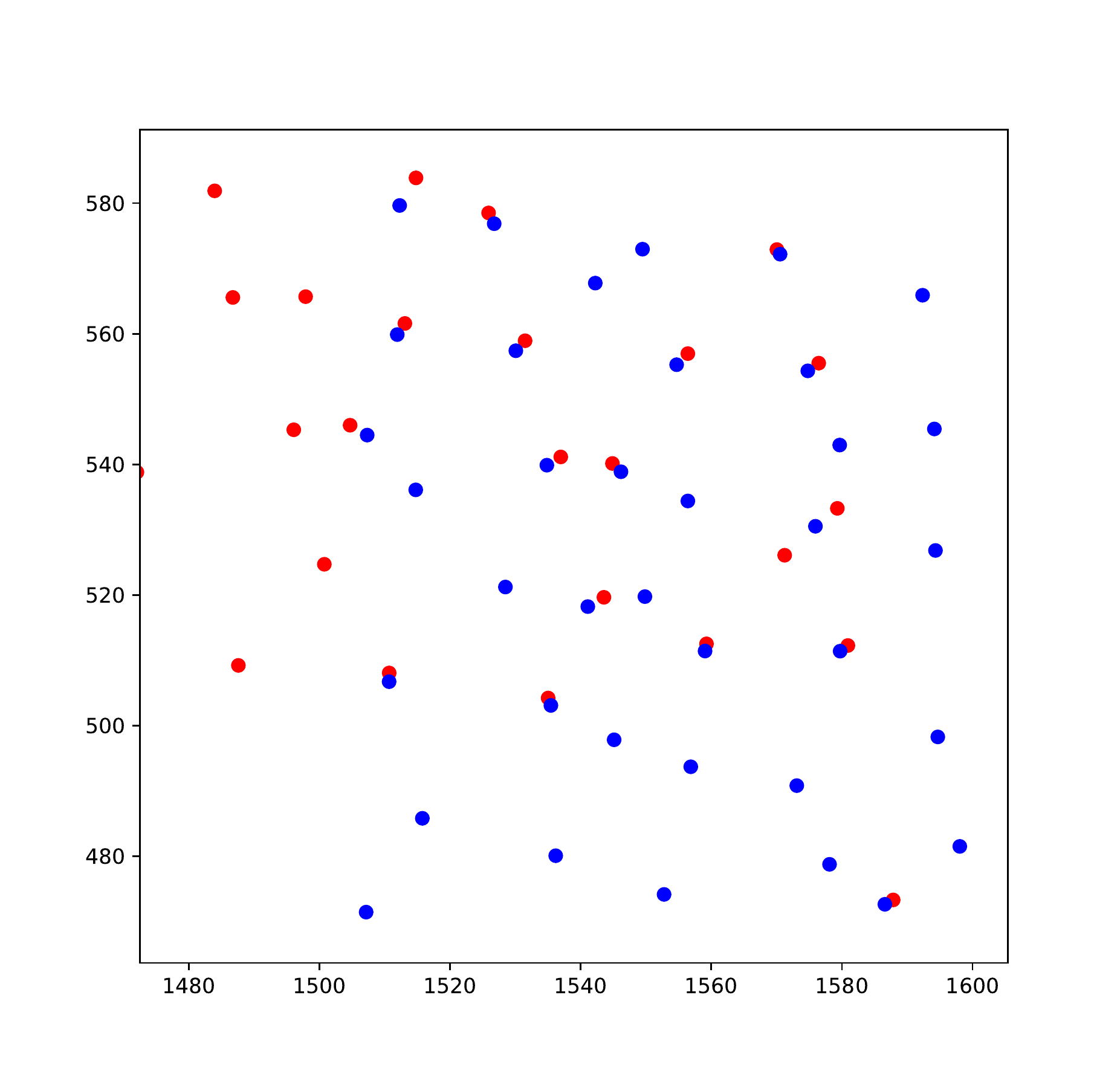}
    \caption{clipping of \figref{fig:defocused laser simple combined} of the top overlapping region}
  \end{subfigure}\\
  \begin{subfigure}[t]{0.7\textwidth}
    \includegraphics[width=\textwidth, trim={5cm 3cm 5cm 4cm}, clip]{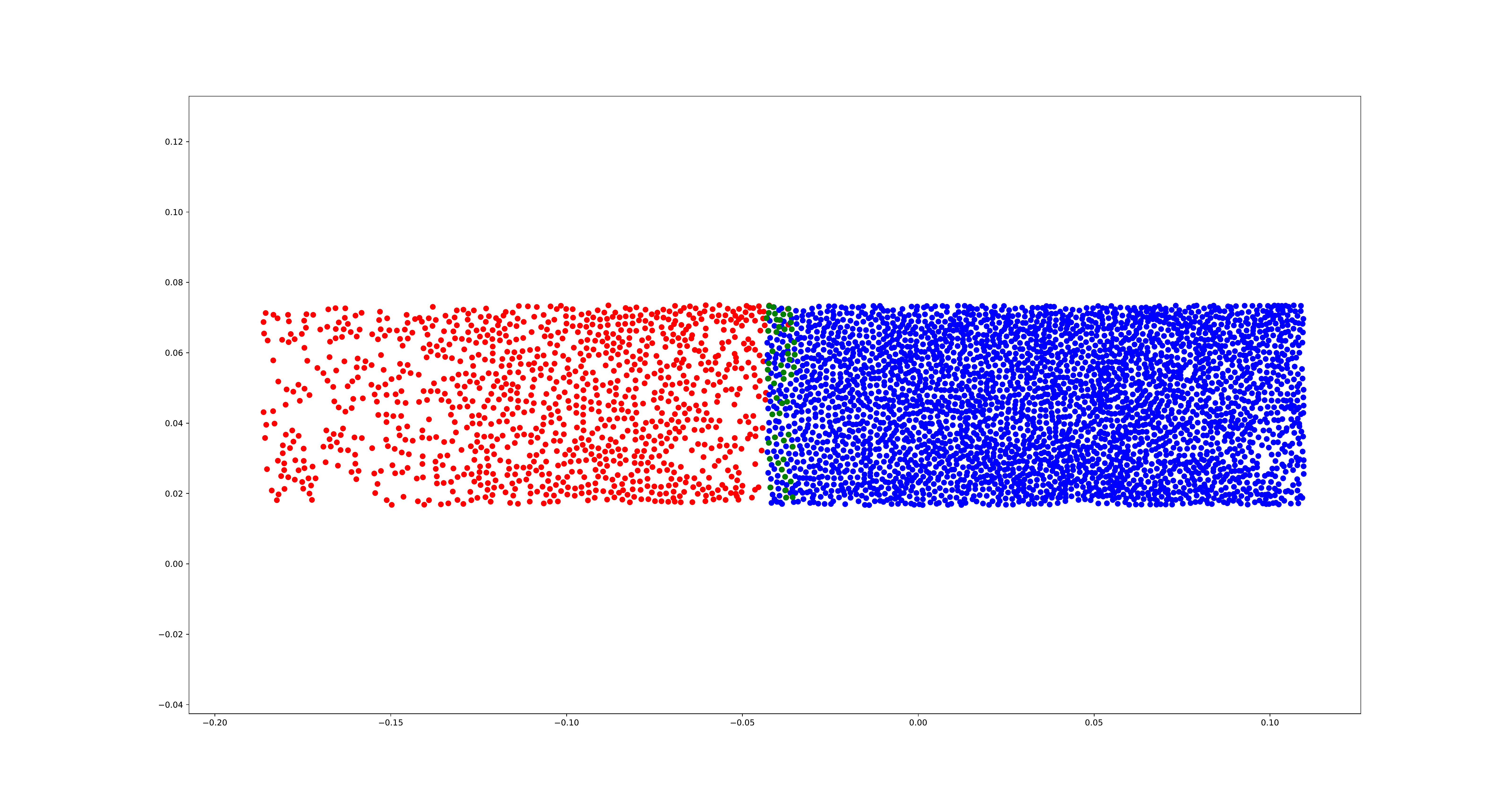} % 650 mm x 345 mm
    \caption{combined features}\label{fig:defocused laser combined}
  \end{subfigure}
  \begin{subfigure}[t]{0.29\textwidth}
    \includegraphics[width=\textwidth, trim={2.5cm 2.5cm 2cm 2.5cm}, clip]{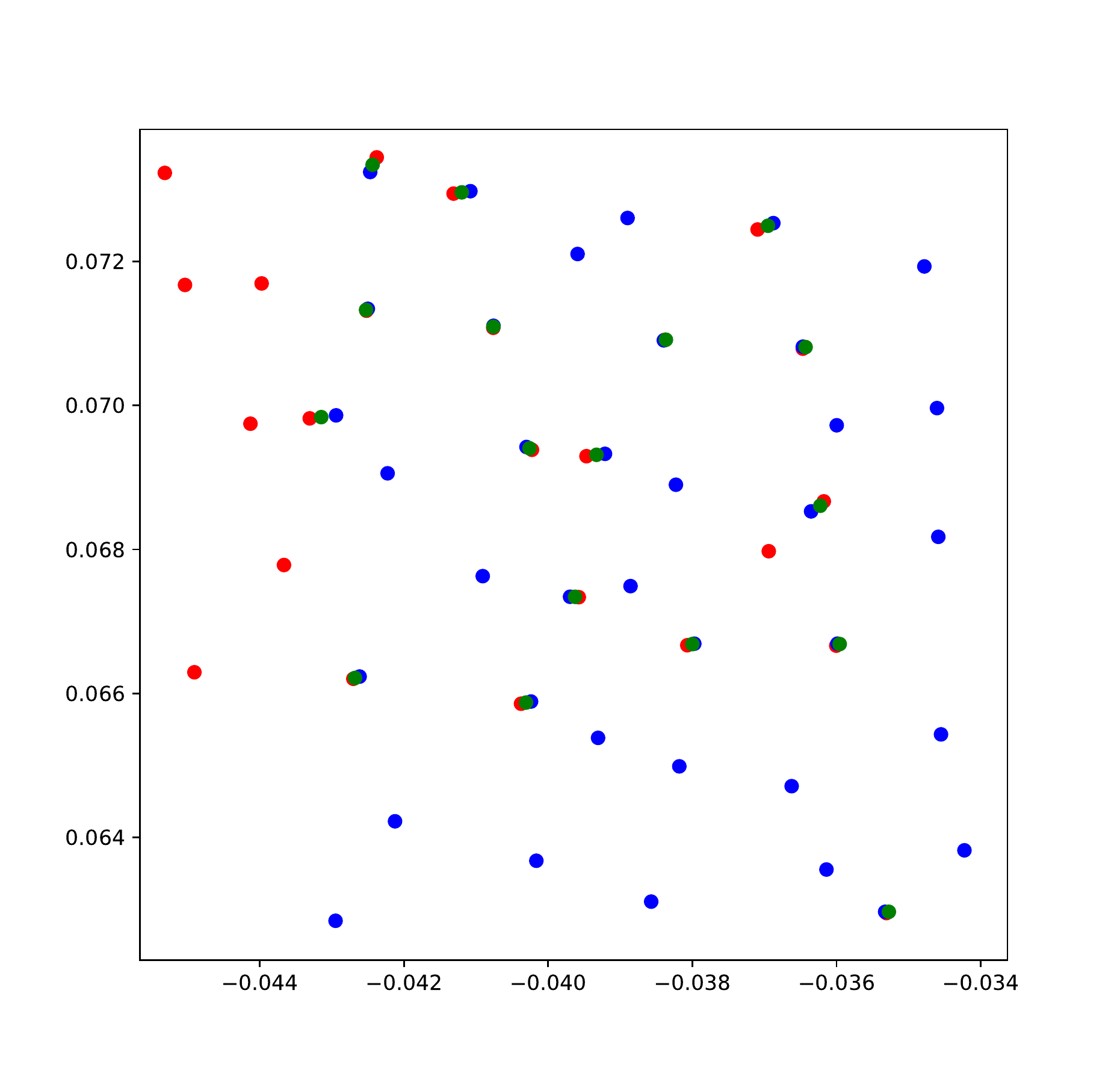}
    \caption{clipping of \figref{fig:defocused laser combined} of the top overlapping region}\label{fig:clipping defocused laser combined}
  \end{subfigure}
  \caption{Projection to laser plane: Example with defocused laser}\label{fig:projection laser plane defocused laser}
\end{figure}

To solve the optimization problem we use a least square solver from scipy \citep{scipy}.

In \figref{fig:projection laser plane focused laser} and \figref{fig:projection laser plane defocused laser} we can see examples of projecting the features of two cameras to the laser plane. We can see that the projection seems to work for most particles. But for one in the clipping in \figref{fig:clipping defocused laser combined} the matching is not perfect. The reason is the position in the depth --- the particle is not located in the assumed zero-width laser plane.

\subsection{Projection to Laser Sheet}
The laser plane can be extended to a laser sheet with a depth described by the parameters $l_l \in \R$, $l \in \R^3$ and $l_u \in \R$:
  
\begin{align*}
  s_i \left(\begin{array}{c}
    p_j^{(i)} \\
    1 \\
  \end{array}\right) = P_i \left(\begin{array}{c}
    x_j \\
    1 \\
  \end{array}\right) \mbox{ for all } x_j \in F_{oV} \subset \R^3
\end{align*}

\begin{align*}
  \forall x_j \in F_{oV}: l_l \leq l^T x_j \leq l_u
\end{align*}

For the example shown in \figref{fig:projection laser plane focused laser} we get $1321$ particles in left camera and $259$ particles in right camera with $48$ corresponding features in the overlapping area. This leads to $274$ unknowns and $288$ equations.

Again we have to find the best solution (minimized reprojection error) by solving the least square optimization problem:
\begin{align*}
  \mbox{obj. func.: } & \sum_{j=1}^{n}{\sum_{i=1}^{m}\norm{
  \left(\begin{array}{c}
    p_j^{(i)} \\
    1 \\
  \end{array}\right) - \frac{1}{s_j^{(i)}} P_i \left(\begin{array}{c}
    x_j \\
    1 \\
  \end{array}\right)
    }^2} \rightarrow \min_{s_j^{(i)}, P_i, x_j}\\
  \mbox{ s.t. } & \forall x_j:  \frac{-d}{2} \leq x_j^T \left(\begin{array}{c} 0 \\ 0 \\ 1 \\ \end{array}\right) \leq \frac{d}{2}
\end{align*}
The number of scalar unknowns and number of residuals of the least square problem is given in \tabref{tabular:equations features unknowns laser with depth}.
\begin{table}\center
  \begin{tabular}{c|c|c|c}
    number of & number of corresponding & number of  & number of \\
    cameras & pairs of features & scalar unknowns & equations \\\hline
    $m$ & $n$ & $n m + 9m + 3n$ & $3 n m$ \\\hline
    $2$ & $18$ & $108$ & $108$ \\\hline
    $3$ & $9$ & $81$ & $81$ \\\hline
    $4$ & $8$ & $92$ & $96$ \\\hline
    $5$ & $7$ & $101$ & $105$ \\
  \end{tabular}
  \caption{This table shows the number of scalar unknowns and equations (number of residuals of the least square problem) as a function of the number of corresponding pairs of features and the number of cameras. Here it is assumed $f_x = f_y$ and the laser sheet is in the z plane ($\frac{-d}{2} \leq \left(0, 0, 1\right) x_j \leq \frac{d}{2}$).}\label{tabular:equations features unknowns laser with depth}
\end{table}

\begin{figure}\center
  \begin{subfigure}[t]{0.45\textwidth}
    \includegraphics[width=\textwidth, trim=3cm 1cm 1.5cm 1.5cm, clip]{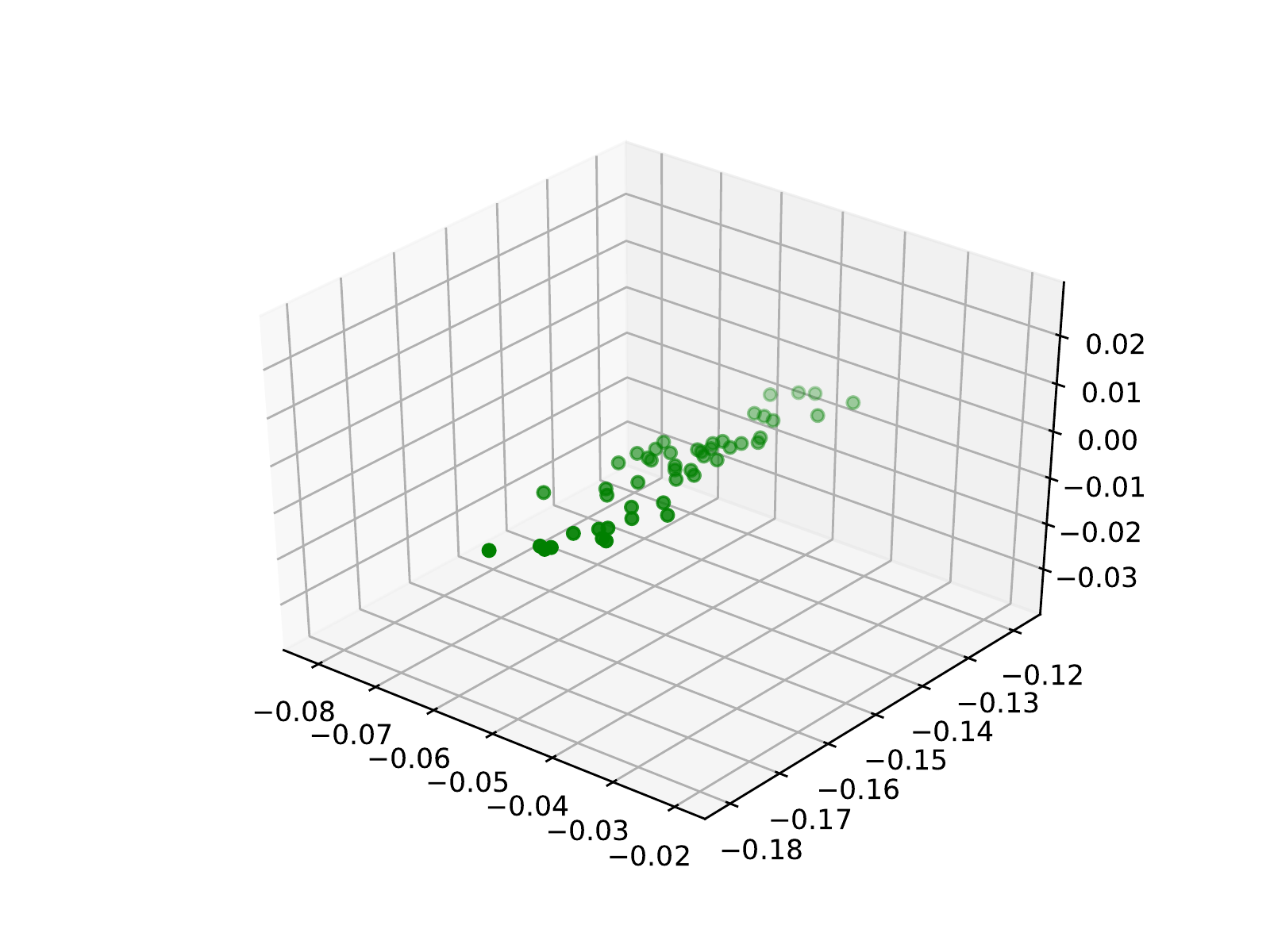}
    \caption{focused laser}
  \end{subfigure}
  \begin{subfigure}[t]{0.45\textwidth}
    \includegraphics[width=\textwidth, trim=3cm 1cm 1.5cm 1.5cm, clip]{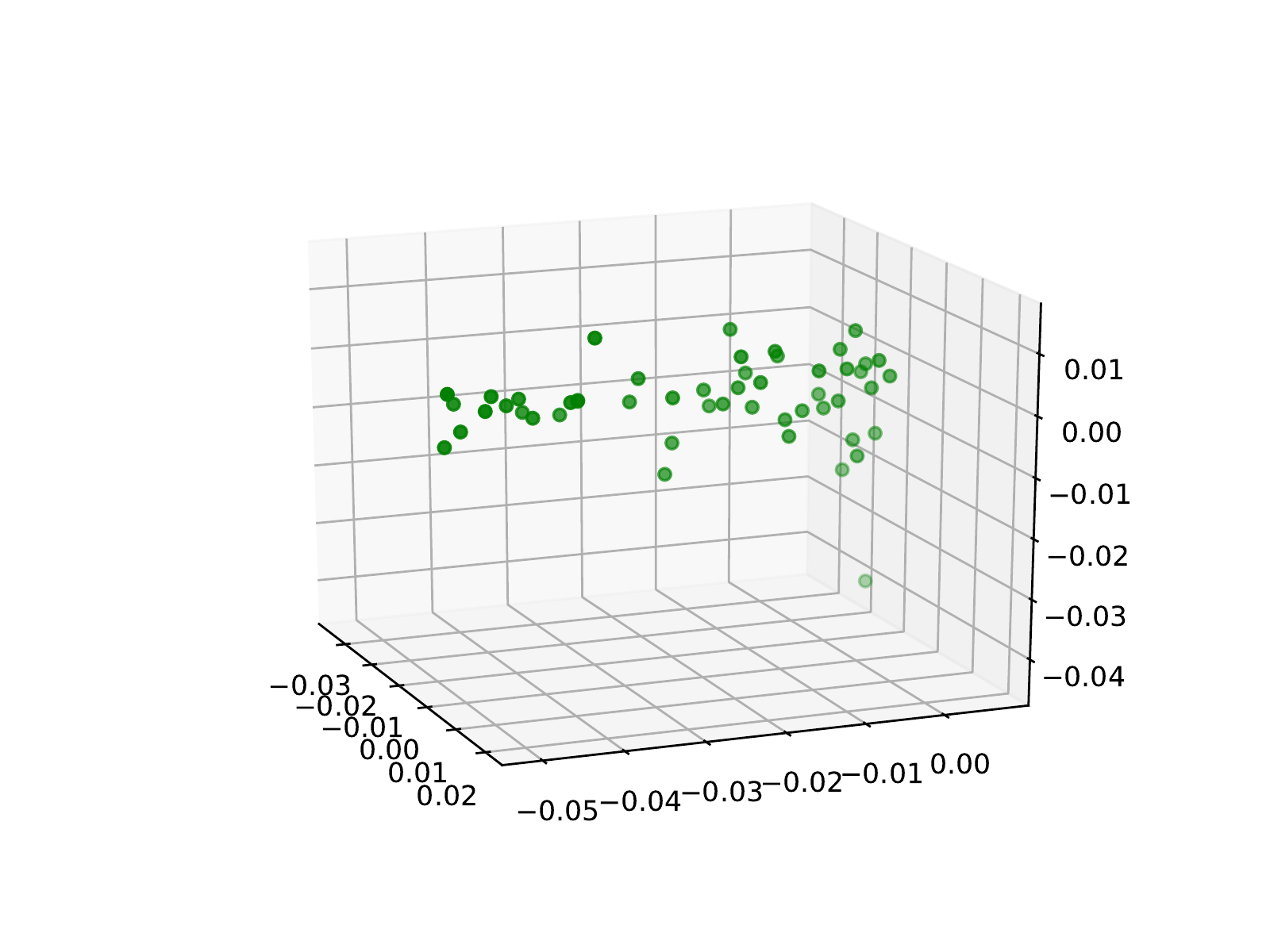}
    \caption{defocused laser}
  \end{subfigure}
  \caption{3D coordinates}\label{fig:3D coordinates}
\end{figure}

In \figref{fig:3D coordinates} two examples of 3D coordinates are shown, which were determined by solving the optimization problem using a least square solver from scipy \citep{scipy}.

\section{Conclusion}
It seems to be possible to get 3D coordinates (\figref{fig:3D coordinates}) just from the images of the PK-4 setup \citep{Pustylnik:2016} in arbitrary units. The resolution approximated in \secref{sec:Possible Accuracy in View Direction} can be enhanced by using subpixel resolution as described by \citet{mohr:2019}. Using the information from \citet{Pustylnik:2016} that $1 \mbox{ px}$ corresponds to about $14 \mbox{ $\mu$m}$ allows to calibrate the arbitrary units with physical units.

\bibliographystyle{named}
\bibliography{literature}\addcontentsline{toc}{section}{\refname}

\end{document}